# Human Activity Recognition from Smartphone Sensor Data for Clinical Trials

Stefania Russo,[1*] Rafał Klimas,[1] Marta Płonka,[2] Hugo Le Gall,[1] Sven Holm,[1] Dimitar Stanev,[1] Florian Lipsmeier[1], Mattia Zanon,[1] Lito Kriara[1*]

[*]Contributed equally

[1]F. Hoffmann-La Roche Ltd, Basel, Switzerland

[2]Roche Polska Sp. z o.o., Warsaw, Poland

**Corresponding author**

Lito Kriara

F. Hoffmann-La Roche Ltd, Basel, Switzerland

lito.kriara@roche.com

# Abstract

We developed a ResNet-based human activity recognition (HAR) model with minimal overhead to detect gait versus non-gait activities and everyday activities (walking, running, stairs, standing, sitting, lying, sit-to-stand transitions). The model was trained and evaluated using smartphone sensor data from adult healthy controls (HC) and people with multiple sclerosis (PwMS) with Expanded Disability Status Scale (EDSS) scores between 0.0–6.5. Datasets included the GaitLab study (ISRCTN15993728), an internal Roche dataset, and publicly available data sources (training only). Data from 34 HC and 68 PwMS (mean [SD] EDSS: 4.7 [1.5]) were included in the evaluation. The HAR model showed 98.4% and 99.6% accuracy in detecting gait versus non-gait activities in the GaitLab and Roche datasets, respectively, similar to a comparative state-of-the-art ResNet model (99.3% and 99.4%). For everyday activities, the proposed model not only demonstrated higher accuracy than the state-of-the-art model (96.2% vs 91.9%; internal Roche dataset) but also maintained high performance across 9 smartphone wear locations (handbag, shopping bag, crossbody bag, backpack, hoodie pocket, coat/jacket pocket, hand, neck, belt), outperforming the state-of-the-art model by 2.8% – 9.0%. In conclusion, the proposed HAR model accurately detects everyday activities and shows high robustness to various smartphone wear locations, demonstrating its practical applicability.

## Introduction

Recent advancements in digital health technologies have demonstrated the feasibility of assessing functional disability related to neurologic disorders such as multiple sclerosis[1-5], Parkinson's disease[4,6], and Huntington's disease[7]. For instance, digital measures of gait can be obtained with specialized algorithms applied to time series of inertial measurements unit (IMU) data collected with consumer-grade smartphones[8-10]. Such digital measures have shown good test-retest reliability[1,5,9-11], agreement with standard clinical measures of gait and disease severity[1,5,10,12], ability to distinguish between patients with mild versus moderate disability[8], and sensitivity to longitudinal changes in disease trajectories[13].

Unlike clinician-administered tests, digital tests can be remotely self-administered at home without supervision by a healthcare professional[1,2], making decentralized trials a reality[2]. Such trials are less disruptive to the patient and may improve access to study participation. Thus, digital tests have potential to reach a wider, more diverse patient population[14] and, consequently, may help enroll demographic groups that have been typically underrepresented in clinical trials. Additionally, digital tests have been shown to offer more sensitive and more objective measures than clinician-administered tests, thereby allowing for shorter, smaller clinical trials[15]. This is particularly important for rare diseases, where recruiting sufficiently large study cohorts for critical phase 2 trials may be challenging.

To capitalize on these advantages, it is crucial that the study participants keep completing the digital tests on a regular basis. However, adherence to these tests remains a challenge[16]. While supervision provided in a clinical trial may improve adherence[2], in longer trials, a drop in adherence can, nonetheless, be observed over time[17]. Hence, new ways of assessing gait with smartphone sensors are needed.

One possible solution that has been explored in recent studies is passive monitoring[18,19]. Unlike active tests, passive monitoring requires no input or interaction by the patient. Instead, sensor data (IMUs) are passively recorded while the patient goes about their daily routine while carrying with them or wearing the measurement device[3,20]. As a result, passive monitoring can generate substantially more data than active tests[2]. Once such data have been collected, a human activity recognition (HAR) model is used to identify bouts of everyday activities such as walking, turning, climbing or descending stairs, standing up or sitting down, sitting, or standing[20-22]. Digital measures extracted from such activity bouts may provide complementary clinical insights on the daily functioning of the patient[3,20,23].

Several different HAR models are publicly available to date[24-26]. Although these models typically perform well on data that are comparable to the data they were initially trained on, they often require substantial retraining efforts before applying them to dissimilar data, which limits their out-of-the-box use. To address this gap, we have developed a HAR model that requires only minimal retraining, is easily adaptable to new data modalities and patient populations and is robust to different sensor wear locations. Here, we evaluate the accuracy of our HAR model in classifying different activities, compare it against the state-of-the-art model, and assess its robustness to multiple sclerosis.

## Methods

### Data sources

Data from different data sources containing smartphone sensor data of everyday activities were used to train, validate, and evaluate the HAR model. These included the in-clinic GaitLab dataset[27], the internal Roche dataset, as well as publicly available data sources[28-35]. Table 1 summarizes the different cohorts enrolled for each of these datasets as well as the activities they performed and the devices they used.

**Table 1.** Data sources used for the training, validation and testing of the proposed HAR model. Additional information (i.e., cohort, activities, number of simultaneous devices, device locations).

| Data sources | Study cohort | Gait activities | Non-gait activities | Devices and device locations |
|---|---|---|---|---|
| GaitLab (in-lab dataset)[27] | • Healthy controls<br>• People living with MS | • Structured in-lab treadmill 2MWT<br>  – Fixed slow place of 2km/h<br>  – Comfortable, self-selected pace<br>  – Self-selected fast pace<br>  – Comfortable self-selected pace while performing a cognitive task<br>• Structured in-lab overground 2MWT<br>  – As fast as possible, but safely<br>• T25FW<br>• UTT | • Static Balance Test battery (standing)<br>  – Natural stance, eyes open<br>  – Natural stance, eyes closed<br>  – Full tandem stance (left leg in front), eyes open<br>  – Full tandem stance (right leg in front), eyes open<br>  – Left leg balance, eyes open<br>  – Right leg balance, eyes closed<br>  – Parallel stance, eyes open | • 6 different smartphones<br>  – Belt, front<br>  – Belt, back<br>  – Pocket, front left<br>  – Trouser pocket, front right<br>  – Trouser pocket, back left<br>  – Trouser pocket, back right |
| GaitLab (real-world dataset)[27] | • Healthy controls<br>• People living with MS | • Unstructured real-world walking[a] | • Not applicable | • 1 singe smartphone<br>  – Belt, front |
| Internal Roche dataset | • Healthy controls | • Walking<br>  – Straight line<br>  – Freestyle<br>  – U-turn<br>• Running<br>• Climbing stairs<br>  – Climbing stairs up<br>  – Climbing stairs down | • Sitting<br>• Sit-to-stand transitions (standing up and sitting down)<br>• Standing<br>  – With arms kept still<br>  – With 'active' arms (ie, moving arms mimicking using a smartphone or gesticulating) | • 9 different smartphones<br>  – Handbag<br>  – Shopping back worn over the shoulder<br>  – Crossbody bag<br>  – Backpack<br>  – Hoodie pocket<br>  – Coat/ jacket pocket<br>  – Hand<br>  – Hanging around the neck<br>  – Belt, back |

| Publicly available data sources[28-35] | • Healthy controls | • Walking<br>• Running<br>• Climbing stairs | • Standing<br>• Sitting<br>• Lying down | • 1 single smartphones<br>− Different locations from subject to subject |
|---|---|---|---|---|

[a]Data from the unstructured real-world walking activity were used only in the analysis on the correlation between ambulatory bout duration detected by the HAR model and MS disease severity.
2MWT, Two-Minute Walk Test; MS, multiple sclerosis; T25FW, Timed 25-Foot Walk; UTT, U-Turn Test.

The design of the GaitLab study (ISRCTN15993728) has been previously reported in detail[27]. The analyses presented here use the smartphone sensor data collected during the second of two on-site visits in a gait laboratory from both healthy controls (HC) and people with multiple sclerosis (PwMS). Activities included seven different walking and seven different balance activities. These activities were retrospectively grouped as gait and non-gait activities, respectively. Smartphone sensor data were simultaneously collected from six different smartphone devices, each being placed on a different part of the body. The sampling frequency was 50 Hz. Gait data were also collected during a 10 – 14 day remote testing period. However, these data were only used to assess the association between the duration of ambulatory bouts detected by the HAR model and MS disease severity. The different gait and non-gait activities from the GaitLab dataset are described in Table 1.

In comparison, the internal Roche dataset included additional activities and smartphone wear locations. Activities included walking in a straight line, freestyle walking, running, making U-turns, standing, standing with active arms (i.e., standing while moving the arms), climbing stairs up & down, standing up & sitting down, and sitting. These activities, which lasted for 1 – 2 minutes and were performed by HC in a single, informal session. The individual activities were grouped into either one of three gait activities (walking, running, stairs activities) or one of three non-gait activities (sitting/lying, standing, sit-to-stand). The additional smartphone wear locations (Table 1) were identified on a questionnaire administered in the GaitLab study as one of the more preferred smartphone wear locations (data on file). As in the GaitLab study, the sampling frequency was 50Hz.

The publicly available data sources consisted of smartphone data from the PAMAP2, HAPT, WISDMAr, WISDMAt, REALDISP, HHAR, and MHEALTH datasets[28-35]. These datasets contain smartphone sensor data collected from a single device while HC performed activities such as walking, running, stairs, standing, and sitting/lying. In 80% of the samples, the smartphone was carried in the trouser pocket close to the thigh.

The different durations of gait and non-gait activities led to an imbalance of the amount of gait/non-gait data, with the proportion of gait and non-gait data varying from dataset to dataset (Supplementary Table S1).

## Proposed HAR model

*Architecture*

The proposed HAR model was designed to identify in smartphone sensor data whether a subject was walking, sitting, lying, stairs, sit-to-stand (both standing up and sitting down), or running. Its architecture is illustrated in Figure 1. The model uses 6-second windows of smartphone accelerometer data sampled in the x-, y-, and z-direction (resampled at 20 Hz using cubic spline interpolation) as input; and predicts for the central 2 seconds of each 6-second window the probability of each of the six activities. The central two seconds are subsequently labeled according to the activity with the highest probability.

The underlying deep learning architecture is inspired by the residual network (ResNet) architecture of He *et al.* that was developed for image recognition tasks[36]. ResNets consist of multiple stacked building blocks (BasicBlock) of convolution-ReLU-convolution layers with a skip connection[37]. These skip connections enable gradient information to be passed through the layers, thereby avoiding the vanishing gradient problem[36].

For the proposed HAR model, we generalized the ResNet of He *et al.*[36] to adapt it to smartphone sensor data recorded during various everyday activities. Hence, the proposed model takes an input tensor of shape *(N, $C_{in}$, L)*, where *N* is the number of windows in a batch during batch training, $C_{in}$ is the size of input channels, and *L* is the number of samples in a window. The input tensor is forward-propagated through the network and becomes an output tensor of shape *(N, $C_{out}$)*, where $C_{out}$ is the number of activities the model is able to predict (six activities in our case), and a value $v_{ij}$ in the tensor represents the probability that the model predicts window *i* should be labelled as activity *j*. Furthermore, the proposed HAR model uses for each of the three BasicBlocks a one-dimensional dilated convolution layers rather than a two-dimensional convolution layers.

The proposed HAR model takes advantage of batch normalization to both reduce the time needed for training and to increase the stability of the training. This normalization is achieved by recentering and rescaling. Additionally, average pooling is applied to reduce the spatial dimensions of feature maps by computing the average of local regions, which helps preserve feature information while reducing complexity and maintaining translation invariance. This is particularly useful in the final global average pooling layer to generate compact feature representations. Subsequently, softmax converts the raw output scores in the final layer to probability through normalizing the exponential of each output by the sum of all exponentials.

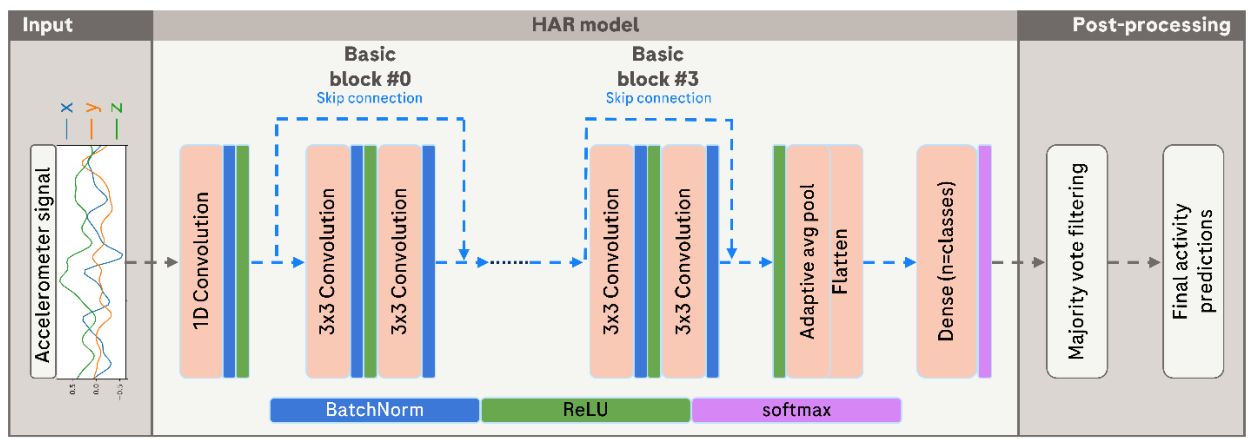

**Figure 1.** The proposed HAR model uses a one-dimensional ResNet architecture with one-dimensional convolutional layers and 3×3 kernels to predict activities from accelerometer signals. Post-processing steps are detailed in Figure 2. Avg Pool, adaptive average pooling; BatchNorm, batch normalization; ReLU, rectified linear unit.

*Data preparation*

As smartphones can be carried in different orientations, an ideal HAR model must be robust towards different smartphone rotations. Data augmentation was, therefore, introduced in the training phase to increase the robustness to smartphone orientation, with the raw smartphone accelerometer data randomly rotated between 0 – 180 degrees around the x-, y-, and z-axis.

Next, the sensor data from all three data sources were split into a training, validation, and evaluation set as described below. This split was done on a participant level to avoid that data from the same participant were being used by multiple sets. The only exception to this was the internal Roche dataset due to its smaller sample size as described below. The number of samples in each of these three sets are reported in Supplementary Table S2.

*Training*

Training the proposed HAR model used a one-cycle learning rate scheduler[38] that was implemented in PyTorch[39]. The maximum learning rate was 0.001. At each iteration (or "epoch"), the input data were randomly bundled as batches of 64 windows (N = 64 in the model architecture section above) and passed through the model. The model subsequently performed convolutional operations on the batches based on the weights defined in the layers and outputted the predicted values. A cross entropy loss[40] was calculated comparing the predicted values with the true activity labels. The weights of the layers in the model were subsequently updated based on a stochastic gradient descent algorithm[41]. After each epoch, the model predicted the activity on data from the validation set and calculated a validation loss on the validation samples. The model was trained for 350 epochs, and model weights with the minimal validation loss were picked for the final model for subsequent evaluation.

## Post-processing

Erroneous HAR model predictions can lead to very short activity bouts (e.g., 2-second bouts) that are not necessarily meaningful. By smoothing the individual HAR model predictions, the robustness of the model's output can be increased. This was achieved through majority vote filtering applied to the HAR model's output[42-44]. With this filtering, the central two seconds of a 6-second sliding window were assigned the predicted activity in the 6-second window that had the highest confidence (i.e., the probability calculated by the HAR) (Figure 2).

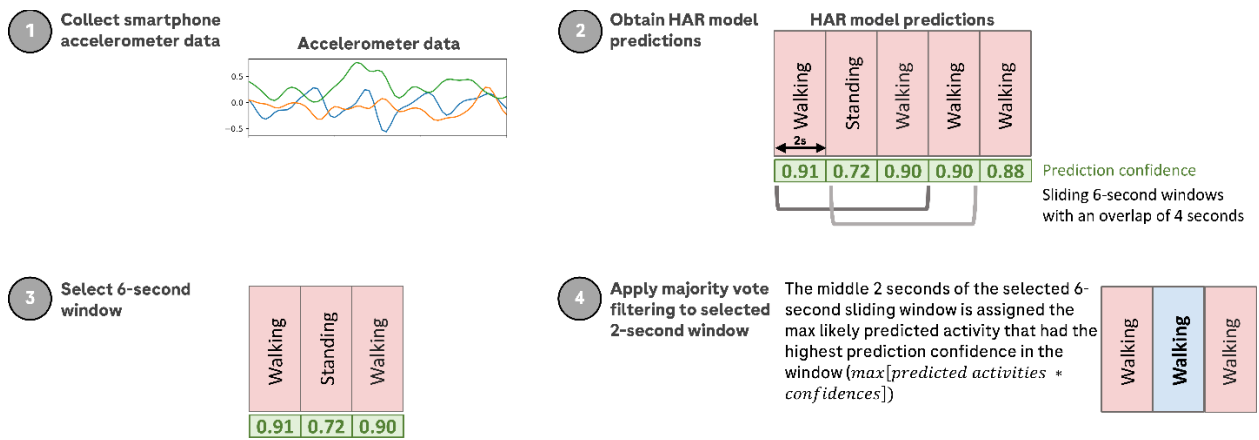

**Figure 2.** Majority vote filtering is applied to ensure the extracted ambulatory bouts are stable and meaningful.

## State-of-the-art HAR model

The state-of-the-art HAR model is the model of Yuan *et al.*[45], which is based on the convolutional neural network architecture of He *et al.*[46] The state-of-the-art HAR model differed from the proposed HAR model in that was trained using a self-supervised approach and required supervised fine-tuning. During the training, the model learnt to predict different clusters. The models was trained on approximately 6 billion smartwatch samples, split into 10-second windows. These samples were recorded at a sampling frequency of 30 Hz from approximately 100,000 participants included in the UK biobank[47]. We fined-tuned this model with data from both the GaitLab study and the internal Roche dataset to both adapt the model to smartphone sensor data and to assign the predicted clusters to one of the everyday activities present in the GaitLab and internal Roche dataset. For the fine-tuning, the smartphone accelerometer data were down-sampled to 30 Hz using cubic spline interpolation and split into 10-second windows, thus matching the data used for the self-supervised training. The fine-tuning was conducted separately for the GaitLab and internal Roche dataset.

## Statistical analysis

### Evaluation of the HAR model's performance

Our HAR model was evaluated against the state-of-the-art HAR model using the evaluation sets of the GaitLab study and the Internal Roche dataset. Given the different number of samples contained in each evaluation set, cross-validation was used to assess the performance of the HAR model on the GaitLab dataset, and leave-one-out cross-validation to assess the model's performance on the Internal Roche dataset. Performance of the HAR model was evaluated by computing accuracy (i.e. the proportion of correctly classified samples), precision (i.e. positive predictive value), recall (i.e. sensitivity), and the F1-score (i.e. the harmonic mean of precision and recall).

### Detection of gait versus non-gait

The ability of detecting gait versus non-gait is critical for extracting gait-related digital measures from passively acquired sensor data. Hence, this analysis evaluated the performance of both HAR models in classifying gait and non-gait activities using data from the GaitLab and internal Roche datasets. The activities detected by the HAR models were post-hoc assigned as 'gait' or 'non-gait'. Subsequently, accuracy (ie, overall accuracy), precision, recall, and F1-score of detecting gait and non-gait activities were computed across all samples collected from all study participants and smartphone wear locations. Additionally, the impact of different disease severity bands on the classification accuracy (PwMS with Expanded Disability Status Scale [EDSS] = 0 – 3.5 vs PwMS with EDSS = 4.5 – 5.5 versus PwMS with EDSS = 6.0 – 6.5) on accuracy of detecting gait and non-gait activities was evaluated on the GaitLab's evaluation set, whereas the impact of smartphone wear locations was studied using data from both GaitLab and internal Roche datasets.

### Detection of everyday activities

Detecting specific everyday activities may provide a more granular picture of the disease[20]. This analysis evaluated the performance of both HAR models in detecting, or classifying, the different activities included in the GaitLab and internal Roche datasets (Table 1). Overall accuracy, precision, recall, and F1-score were computed across samples (ie, across all activities and smartphone wear locations). Additionally, the accuracy of correctly detecting everyday activities was computed for each activity (across all smartphone wear locations) and for the different smartphone wear locations (across all activities).

*Duration of ambulatory bouts detected by the proposed HAR model versus MS disease severity*

This analysis evaluated whether the mean ambulatory bout duration during passive monitoring is related in the GaitLab study with MS disease severity. To collect passive monitoring data, PwMS were instructed to carry a single smartphone device in a belt worn at the waist level for at least 15 minutes and up to 4 hours a day during the 2-week remote testing period[27]. All continuous gait predictions were aggregated into individual ambulatory bouts, and subsequently the mean ambulatory bout duration across all ambulatory bouts was computed on a subject level, resulting in one measure of mean ambulatory bout duration per subject. Differences in mean ambulatory bout duration across MS disease severity groups (PwMS with EDSS = 0 – 3.5 vs PwMS with EDSS = 4.5 – 5.5 versus PwMS with EDSS = 6.0 – 6.5) and walking aid status groups (use of walking aid: yes vs no) were assessed for statistical significance by the Mann–Whitney U test. A p-value < 0.05 was considered as statistically significant.

# Results

## Participants

The baseline demographics and disease characteristics of those subjects included in the subsequent analyses are reported in Table 2.

**Table 2.** Baseline demographics and disease characteristics of study participants included in the test set of the GaitLab and internal Roche datasets.

|  | Gaitlab dataset | | | | | | Internal Roche dataset |
| --- | --- | --- | --- | --- | --- | --- | --- |
|  | All (n = 89) | HC (n = 21) | All PwMS (n = 68) | PwMS with EDSS 0.0 – 3.5 (n = 20) | PwMS with EDSS 4.0 – 5.5 (n = 22) | PwMS with EDSS 6.0 – 6.5 (n = 26) | All/ HC (n = 13) |
| Female, n (%) | 55 (62) | 13 (62) | 42 (62) | 12 (60) | 14 (64) | 16 (62) | 7 (53.8) |
| Age, years, mean (SD) | 51.9 (13.5) | 52.1 (14.0) | 51.8 (13.3) | 40.8 (9.2) | 53.2 (12.9) | 59.0 (10.6) | 35.8 (4.1) |
| MS phenotype |  |  |  |  |  |  |  |
|   Relapsing-remitting, n (%) | – | – | 47 (69) | 20 (100) | 17 (77) | 10 (38) | – |
|   Secondary progressive, n (%) | – | – | 15 (22) | 0 (0) | 3 (14) | 12 (46) | – |
|   Primary progressive, n (%) | – | – | 6 (9) | 0 (0) | 2 (9) | 4 (15) | – |
| EDDS, mean (SD) | – | – | 4.7 (1.5) | 2.6 (0.5) | 4.7 (0.7) | 6.2 (0.2) | – |
| Ambulation, mean (SD) | – | – | 2.7 (2.3) | 0.4 (0.5) | 1.9 (1.6) | 5.2 (0.8) | – |
| T25FW, s, mean (SD) | – | 3.8 (0.6) | 7.9 (6.9) | 4.1 (0.6) | 5.4 (1.6) | 12.9 (8.6) | – |

EDSS, Expanded Disability Status Scale; MS, multiple sclerosis; T25FW, Timed 25-Foot Walk.

## Detection of gait versus non-gait

The detection of gait versus non-gait activities was evaluated for the proposed HAR model on both the GaitLab and the internal Roche dataset (Figure 3). On the GaitLab dataset, the overall accuracy of correctly classifying gait and non-gait activities was 98.4%, with F1-score, precision, and recall metrics being ≥ 97.9% (Table 3).

Next, we evaluated the proposed HAR model's performance across different levels of disease burden. The detection of gait activities remained high across cohorts (Figure 3a), including HC (mean [SD] accuracy across subjects: 99.5% [0.7%]), PwMS with EDSS score between 0.0 – 3.5 (mean [SD] accuracy across subjects: 99.6% [0.6%]), and PwMS with EDSS score between 4.0 – 5.5 (mean [SD] accuracy across subjects: 98.9% [2.5%]). However, a slight drop in performance was observed in PwMS with EDSS score between 6.0 – 6.5, but the model still performed on a high level (mean [SD] accuracy across subjects: 93.8% [9.1%]). The use of walking aids might have led to improved gait patterns in this subgroup of PwMS, resulting in a more accurate detection of gait activities (mean [SD] accuracy in PwMS with EDSS score between 6.0 – 6.56 who used vs who did not use walking aids: 95.1% [0.08%] vs 84.6% [0.14%]). However, only 3 PwMS in this cohort did not use walking aids (Supplementary Table S3). The model's performance in detecting non-gait activities was consistently high across all cohorts (mean [SD] accuracy across HC: 99.9% [0.2%], across PwMS with EDSS score between 0.0 – 3.5: 99.9% [0.1%], across PwMS with EDSS score between 4.0 – 5.5: 99.6% [0.9%], across PwMS with EDSS score between 6.0 – 6.5: 99.8% [0.4%]; Figure 3a).

**Table 3.** Accuracy in detecting gait versus non-gait in both GaitLab and Internal Roche datasets.

| Metric | GaitLab dataset | | Internal Roche dataset | |
|---|---|---|---|---|
| | **Proposed HAR model** | **Yuan et al.'s model** | **Proposed HAR model** | **Yuan et al.'s model** |
| Overall accuracy | 98.4% | 99.3% | 99.6% | 99.4% |
| F1-score | 98.9% | 97.5% | 99.7% | 99.6% |
| Precision | 99.9% | 95.8% | 99.6% | 99.8% |
| Recall | 97.9% | 99.4% | 99.8% | 99.4% |

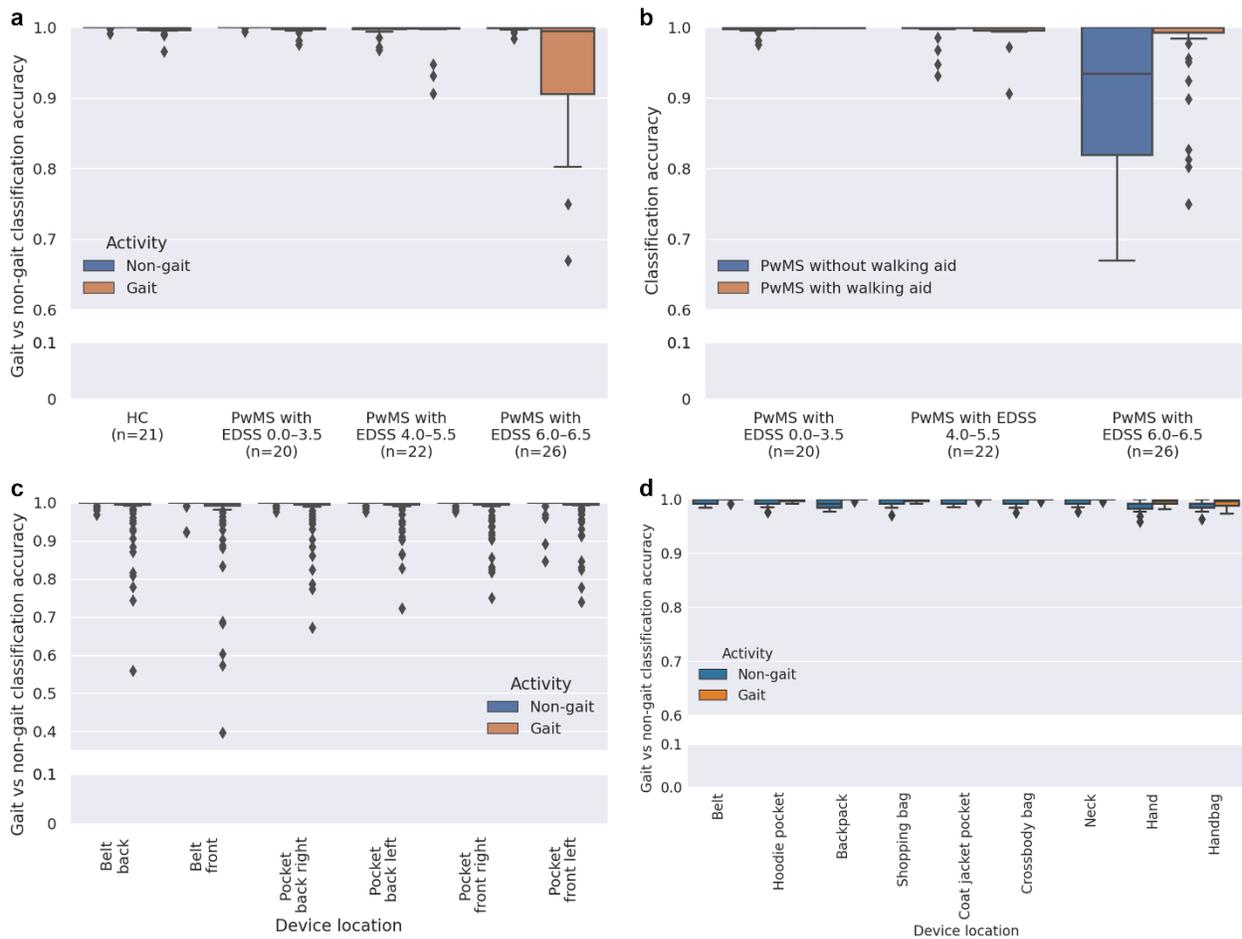

**Figure 3.** Accuracy in detecting gait versus non-gait activities using data from the GaitLab evaluation set (a-c) and Internal Roche dataset (d). (a) The average accuracy in detecting gait and non-gait is comparable across different levels of disease severity except for the detection of gait activities in PwMS with EDSS 6.0 – 6.5. (b) The reduced accuracy in detecting gait activities in PwMS with EDSS 6.0 – 6.5 is strongest among those reporting the use of walking aids. Sample size for each subgroup is reported in Supplementary Table S3. (c) The accuracy in detecting gait and non-gait activities was comparable across smartphone wear locations when using data from the GaitLab dataset (n = 89). (d) The accuracy in detecting gait and non-gait activities was comparable across smartphone wear locations when using data from the internal Roche dataset (n = 13) except for smartphone wear locations characterized by a large degree of movement (i.e., hand and handbag).

On the internal Roche dataset, the proposed HAR model showed an overall accuracy of 99.6% in detecting gait versus non-gait activities across all participants and all smartphone wear locations. F1-score, precision, and recall metrics were also high (all ≥ 99.6%; Table 3).

The HAR model's performance was mostly consistent across the different smartphone wear locations both for the GaitLab (Figure 3c) and the internal Roche (Figure 3d) datasets, with the mean accuracy of detecting gait activities being close to 100% for most locations. However, a small decrease of approximately 1% – 2% in the accuracy was noted when the smartphone was placed in the hand or in a handbag. This is likely due to the larger contribution of arm or upper body movement to the smartphone's accelerometer data.

In both datasets, the performance of our HAR model in detecting gait versus non-gait was similar to that of the state-of-the-art HAR model of Yuan *et al.*[45] (Table 3). The state-of-the-art model, however, had a higher false positive rate (false positive rate: 30%) among PwMS with EDSS score between 6.0 – 6.5 enrolled in the GaitLab study (Supplementary Figure S1). It is possible that PwMS in this subgroup had to either take some steps during the non-gait activities in order to maintain a stable stance, which was detected as gait by the state-of-the-art model, or take more or longer breaks during the gait activities. This could lead to the state-of-the-art model detecting gait during non-gait activities and vice versa, resulting in a higher false positive rate.

### Detection of everyday activities

The GaitLab dataset included only two types of activities: walking and standing. As shown in Figure 4a and 4b, both HAR models detected walking activities as walking activities (≥ 97% accuracy) and standing activities as standing activities with high accuracy (≥ 97% accuracy).

By comparison, the internal Roche dataset enabled us to evaluate the proposed HAR model's performance in classifying six specific everyday activities: running, sitting/lying, stair walking, standing, walking, and sit-to-stand transitions. The accuracy across all participants, smartphone wear locations, and activities was 96.2%. F1-score, precision, and recall were all 96.2% as well (Table 4). More specifically, the accuracy was 99% for running, 98% for both walking and sit-to-stand transitions, 95% for stair walking, 93% for sitting/ lying, and 92% for standing (Figure 4). The lower accuracy for sitting/ lying and standing likely stems from the similarity of the smartphone sensor signal of these two activities; on rare occasions, the proposed HAR model mistook one activity for the other (Figure 4). Furthermore, the HAR model performed consistently well across the different smartphone wear locations (Table 4). The accuracy ranged from 94.5% – 97.4% across the different smartphone wear locations, with the lowest mean accuracy observed when the smartphone was carried in the handbag or in the hand (accuracy: 94.8% and 94.5%, respectively).

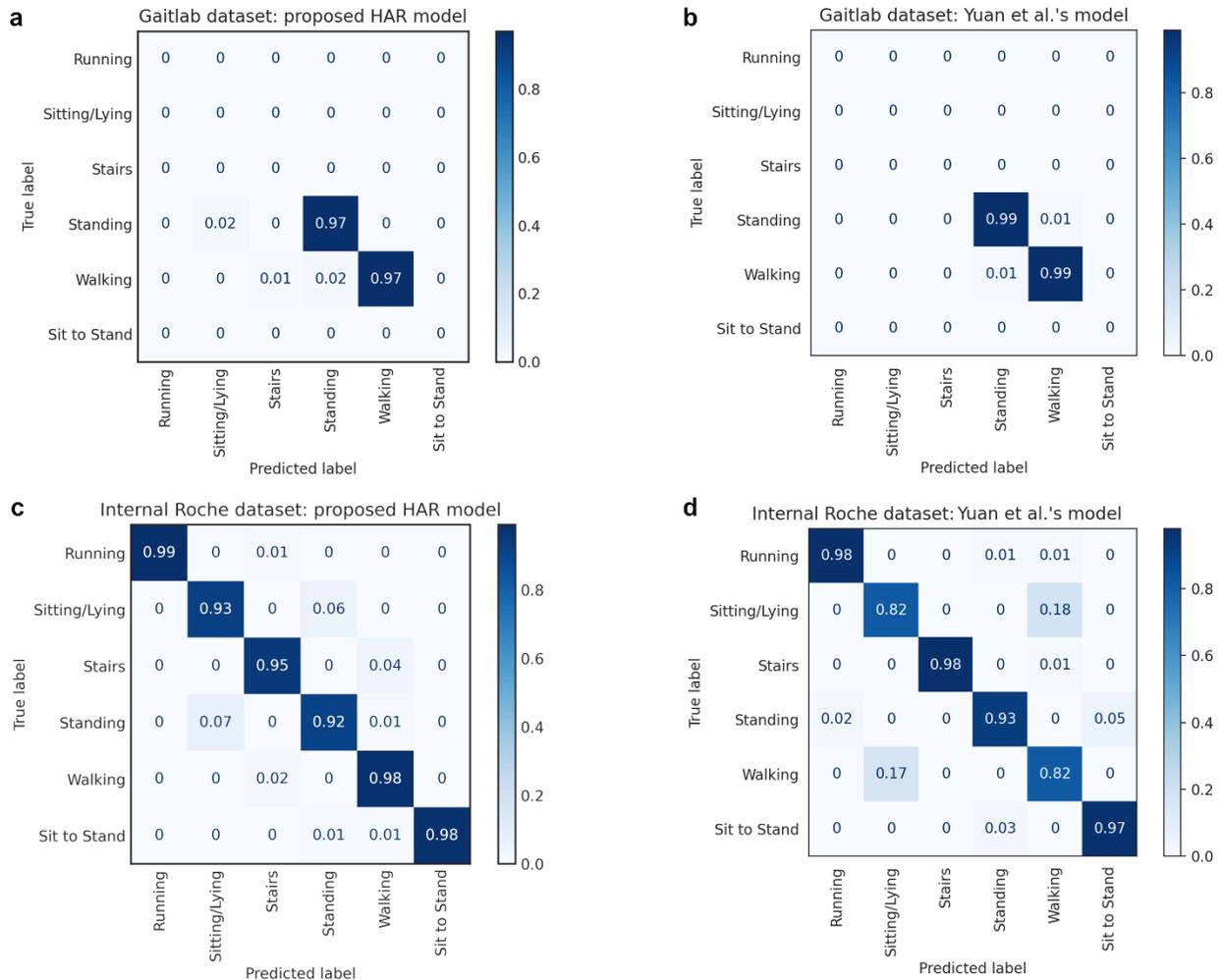

**Figure 4.** Confusion matrix of the HAR's models accuracy in correctly classifying everyday activities (walking, running, stairs, standing, sitting/lying, sit-to-stand) across all smartphone wear locations. On rare occasions, the HAR model may mistake standing for sitting/ lying and vice versa. (a) Confusion matrix of the proposed HAR model applied to the GaitLab dataset. (b) Confusion matrix of the state-of-art model applied to the GaitLab dataset. (c) Confusion matrix of the proposed HAR model applied to the Internal Roche dataset. (d) Confusion matrix for the state-of-the-art HAR model applied to the Internal Roche dataset.

**Table 4.** HAR model performance in classifying everyday activities (walking, running, stairs, standing, sitting/lying, sit to stand).

| Metric | Internal Roche dataset | |
| --- | --- | --- |
| | **Proposed HAR model** | **Yuan et al.'s model** |
| **Overall HAR model performance (across all smartphone wear locations and activities)** | | |
| Overall accuracy | 96.2% | 91.9% |
| F1-score | 96.2% | 92.0% |
| Precision | 96.2% | 92.0% |
| Recall | 96.2% | 91.9% |
| **Accuracy by smartphone wear location (across all activities)** | | |
| Belt | 97.4% | 94.5% |
| Hoodie pocket | 96.3% | 92.0% |
| Backpack | 97.2% | 94.4% |
| Shopping bag | 96.9% | 94.1% |
| Coat/jacket pocket | 97.0% | 92.3% |
| Cross body bag | 96.4% | 92.5% |
| Neck | 95.4% | 86.4% |
| Hand | 94.5% | 90.4% |
| Handbag | 94.8% | 91.1% |

The proposed HAR model's performance in classifying everyday activities is comparable, if not superior, to the performance of the start-or-the-art HAR model of Yuan *et al.*[45] (Table 4). While the accuracy of classifying running, stair walking, standing, and sit-to-stand transitions was comparable between the two models (proposed HAR model vs Yuan et al's model: 92% – 99% vs 93% – 98%), our HAR model outperformed the state-of-the-art model in correctly classifying sitting/lying and walking by more than 10% (sitting/lying: 93% vs 82%; walking: 98% vs 82%, respectively; Figure 4), resulting in an overall improvement in classifying everyday activities by approximately 4% (96.2% vs 91.9%; Table 4). When assessing the average accuracy in detecting these activities by smartphone location, our HAR model outperformed the state-of-the-art model by 2.8% – 9.0%.

## Duration of ambulatory bouts detected by the proposed HAR model versus MS disease severity

Using passive monitoring data collected in the GaitLab study, the mean bout duration of all ambulatory bouts (ie, gait bouts) detected by the proposed HAR model was computed for each study participant. PwMS with EDSS 6.0 – 6.5 had statistically significantly shorter ambulation bout than HC or PwMS with less severe disease (all $p < 0.05$; Figure 5a). The use of walking aids appeared to be associated with a further numerical decrease in ambulatory bout duration (Figure 5b). However, most PwMS with EDSS 6.0 – 6.5 used walking aids during passive monitoring (with vs without walking aid: n = 23 vs n = 1; Supplementary Table S4).

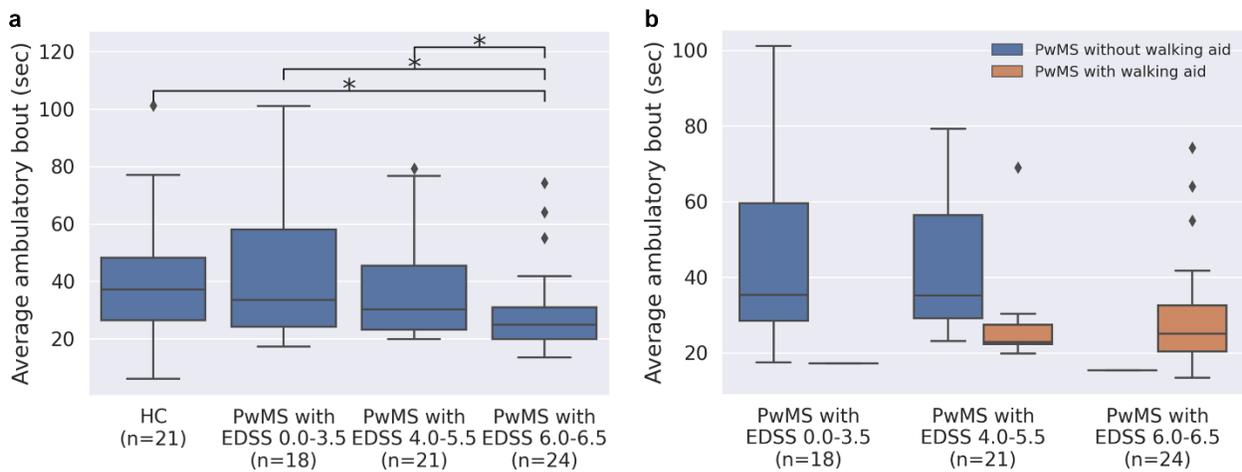

**Figure 5.** (a) PwMS with EDSS 6.0-6.5 show shorter ambulatory bout durations in passive monitoring data detected by the proposed HAR model than HC or PwMS with less severe disease. (b) PwMS using walking aids tend to have numerically shorter ambulatory bouts than PwMS not using walking aids. The sample size of individual subgroups is reported in Supplementary Table S4. *$p < 0.05$ after applying the Mann–Whitney U test. EDSS, Expanded Disability Status Scale; HC, healthy controls; PwMS, people with multiple sclerosis.

## Discussion

We have developed a HAR model that utilizes a ResNet architecture to accurately detect everyday activities using smartphone accelerometer data. The model demonstrates performance in detecting everyday activities that is comparable to, if not better than, a state-of-the-art ResNet model. Remarkably, our proposed HAR model maintains its performance across a wide range of different smartphone wear locations. This is of particular importance for scenarios involving passive monitoring, where study participants or patients are required to carry their smartphones for extended periods. The key learnings form our study are summarized in the panel.

**Panel.** Learnings.

**Is a vast amount of training data required for optimal performance of a HAR model?**

Our HAR model was trained on only 619,444 samples, whereas the model by Yuan *et al.*[45] was initially trained on 6 billion data points. Yet, the proposed HAR model showed higher accuracy in classifying everyday activities such as running, sitting/lying, stair walking, standing, walking, and sit-to-stand transitions(96.2% vs 91.9%). Thus, HAR models trained with limited amount of data can show comparable, if not superior, performance to a HAR model training with vastly more data. Instead, it is more important that the training data is as representative of the data that the HAR model is later being evaluated on.

**Can a HAR model initially trained on data obtained from wrist-worn sensors be fine-tuned to detect activities in smartphone sensor data?**

The HAR model by Yuan *et al.*[45] was initially trained on wrist-worn sensor data. We subsequently fine-tuned this model to accurately detect with high accuracy everyday activities with smartphone data obtained from different smartphone wear locations.

**Does the input signal length affect the HAR model's performance?**

Both our HAR model and that of Yuan *et al.*[45] use the same ResNet architecture. However, our HAR model uses 3-second windows of smartphone sensor data as input, in contrast to the 10-second windows used by Yuan et al.'s model. Despite using a shorter input signal length, our HAR model performed slightly better when detecting the six different everyday activities. For the detection of gait versus non-gait, the difference in the performance of the two HAR models was negligible. Together, this suggests that the input signal length had minimal impact on the HAR model's performance.

However, in the GaitLab dataset, our HAR model has a lower recall than Yuan et al.'s model. This lower recall can be explained by a higher false positive rate in people with severe MS. This may indicate that longer input signals may be beneficial for detecting slow, or more impaired, gait versus non-gait.

**Is it necessary that the training data includes data from PwMS if the HAR model is subsequently applied in MS clinical trials?**

For detecting gait versus non-gait activities, we fine-tuned the HAR model of Yuan *et al.*[45] using either data from HC only or data from both HC and PwMS. In both cases, the overall accuracy of detecting gait versus non-gait was > 98%, indicating high accuracy even if the model is trained on data from HC only.

**Does the smartphone location impact the performance of the proposed HAR model?**

The performance of the proposed HAR model was consistent across most smartphone wear location, thus allowing the study participants to carry the device at their preferred location. A slight drop in performance was only observed when the smartphone was held in the hand or carried in a handbag, possibly due to a greater impact of upper body movement on the smartphone sensor data.

The differences in the performance of the proposed and state-of-the art HAR model of Yuan et al.[45] in detecting gait versus non-gait activities are negligible in most cases. However, the proposed HAR model did not perform as well as the state-of-the-art model on data collected from PwMS with EDSS score between 6.0 – 6.5. This could indicate that the walking patterns in this patient population could be either too slow for our HAR model to optimally detect gait activities or that the gait pattern is sufficiently different in this population, resulting in the HAR model misidentifying certain parts of gait as non-gait. The finding that gait-related algorithms are more erroneous at lower gait speed has been previously reported[48]. Additionally, it is also possible that the longer input signal length used by the model of Yuan *et al.* helps in accurately differentiating between segments of gait versus segments of non-gait (see also Panel).

In contrast, the proposed HAR model outperformed the state-of-the-art model in detecting everyday activities, and in particular sitting or standing/ lying activities. One possible explanation is that the proposed HAR model was partially trained on data from the GaitLab dataset, which contained a significant amount data collected during standing activities. In contrast, the model of Yuan et al. was fine-tuned using data from the Internal Roche dataset only, and was consequently less exposed to standing data.

The proposed HAR model demonstrated its robustness with regards to different smartphone wear locations. This partially stems from leveraging several different data sources for training and validation. Our own data sources—the GaitLab and Internal Roche datasets—offer smartphone accelerometer data from both HC and PwMS performing a wide range of everyday activities. These activities were performed with multiple smartphone devices placed on different parts of the body. Using such data during training made our HAR model less dependent on the smartphone wear location. Additionally, publicly available data sources were leveraged to increase the number of samples used for training. A key advantage of using these data is that they represent a wide range of activities, although data were collected from a single device only.

The proposed HAR model was trained and evaluated on smartphone accelerometer data collected during active tests. These tests provided the study participants with precise instructions on the activity to be performed during the test, and, therefore, provided the true labels needed to train and evaluate the model. HAR models are, however, typically applied to passive monitoring data, where the activities are not known *a priori*. We showed preliminary findings that indicate that PwMS with EDSS score between 6.0 – 6.5 have shorter mean ambulatory bout durations than PwMS with smaller EDSS scores. This finding is in line with previously reported findings that showed that a reduction in life-space—a measure of the geographical space that a person moves through day-to-day—between clinic visits is associated with a concomitant increase in T25FW time.[50,51]

Of note, all ambulatory bouts detected by the proposed HAR model in data collected during the unstructured real-world walking activity were included in the presented analysis. It is, however, conceivable, that short ambulatory bouts of a couple of seconds (eg, from short bursts of walking at home) may not carry the same disease-related signal as longer ambulatory bouts (eg, from walking outdoors).[49], Moreover, computing a larger set digital measures derived from such passively collected gait data may provide additional insights into gait in daily life of PwMS.[3] Thus, the presented analysis is a first step towards utilizing the full potential of the proposed HAR model in the analysis of passively collected smartphone sensor data in MS and other neurologic disorders.

This study comes with several limitations. First, when evaluating the performance of the proposed HAR model, we applied the model exclusively to smartphone accelerometer data collected from pre-specified activities rather than from passive monitoring. However, this allowed us to have the labels for each activity, which was required for training, validating, and evaluating the HAR model. The performance of the proposed HAR in detecting gait, or everyday activities, in passive monitoring data needs to be further evaluated. Second, the internal Roche datasets contained data collected from HC only. Hence, the ability of the proposed HAR model to correctly classify everyday activities could not be assessed in PwMS.

However, the performance of the proposed HAR model in detecting gait versus non-gait was not substantially impacted by MS disease severity; although a slight decrease in the classification accuracy was noted in PwMS with an EDSS score between 6.0 – 6.5. Based on these results, we, therefore, expect that our HAR model performs equally well in classifying everyday activities in most PwMS. To further increase our HAR model's versatility in classifying everyday activities, it would need to be evaluated on data collected not only from PwMS, but also from patients with other neurologic disorders such as Parkinson's or Huntington's disease.

In conclusion, we developed a ResNet-based HAR model trained on a diverse dataset derived from smartphone sensor data. The proposed model demonstrates high accuracy in detecting both gait versus non-gait and in recognizing everyday activities in both HC and PwMS. The model's performance in detecting everyday activities not only matches but potentially exceeds that of the current state-of-the-art HAR model. Additionally, the proposed model's robustness across various smartphone wear locations further enhances its practical applicability. Future research will aim to explore the implementation of this model in passive monitoring analyses, both in MS and other neurologic disorders, thereby extending its utility in remote monitoring of chronic neurologic disorders.

## Data availability

For up-to-date details on Roche's Global Policy on Sharing of Clinical Study Information and how to request access to related clinical study documents, see the website.[52] Request for the data underlying this publication requires a detailed, hypothesis-driven statistical analysis plan that is collaboratively developed by the requestor and company subject matter experts. Such requests should be directed to dbm.datarequest@roche.com for consideration. Anonymized records for individual patients across more than 1 data source external to Roche cannot, and should not, be linked due to a potential increase in the risk of patient reidentification.

## Acknowledgments

We would like to thank the study participants, their families, and the study investigators for the contribution to the studies. We would also like to thank the authors of the PAMAP2, HAPT, WISDMAr, WISDMAt, REALDISP, HHAR, and MHEALTH studies for making their datasets publicly available. The GaitLab and Internal Roche studies were funded by F. Hoffmann-La Roche Ltd, Basel, Switzerland.

## Author contributions

**S Russo** contributed to the methodology, analysis and interpretation of the data; created the software; and drafted and/or critically reviewed and revised the manuscript.

**R Klimas** contributed to the analysis and interpretation of the data; and drafted and/or critically reviewed and revised the manuscript.

**M Płonka** contributed to the analysis and interpretation of the data; and drafted and/or critically reviewed and revised the manuscript.

**S Holm** contributed to the interpretation of the data and drafted and/or critically reviewed and revised the manuscript.

**D Stanev** contributed the design of the studies, as well as the analysis and the interpretation of the data; and drafted and/or critically reviewed and revised the manuscript.

**F Lipsmeier** contributed to the analysis and the interpretation of the data; and drafted and/or critically reviewed and revised the manuscript.

**M Zanon** contributed the design of the studies, as well as the analysis and the interpretation of the data; and drafted and/or critically reviewed and revised the manuscript.

**L Kriara** contributed the design of the studies, the methodology, as well as the analysis and the interpretation of the data; and drafted and/or critically reviewed and revised the manuscript.

## Competing interests

The authors declare the following competing financial interests and no competing non-financial interests:

**S Russo** was a contractor for F. Hoffmann-La Roche Ltd. during the completion of the work related to this manuscript and is now with Takeda Pharmaceutical Co. Ltd., which was not in any way associated with this study.

**R Klimas** is a contractor for F. Hoffmann-La Roche Ltd.

**M Płonka** is a contractor for Roche Polska Sp. z o.o.

Supplementary figures and tables

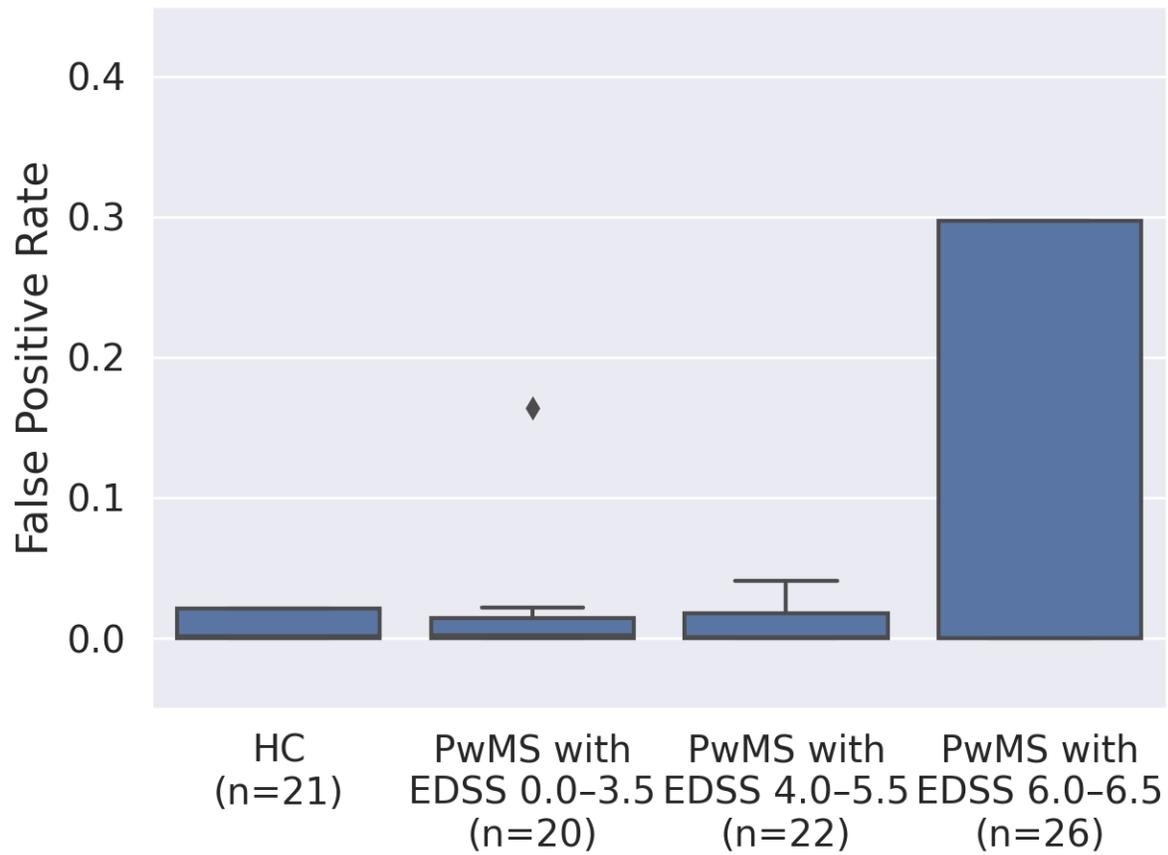

**Supplementary Figure S1.** False positive rates (FPR) of the state-of-the-art HAR model by Yuan et al. [45] applied to the GaitLab test set.

**Supplementary Table S1.** Proportion of gait and non-gait data.

| Dataset | Proportion of gait data, % | Proportion of non-gait data, % |
|---|---|---|
| GaitLab | 75.9 | 24.1 |
| Internal Roche dataset | 64.3 | 35.7 |
| Publicly available data sources | | |
|   PAMAP2Dataset | 49.5 | 50.5 |
|   HAPTDataset | 43.9 | 56.1 |
|   WISDMArDataset | 86.3 | 13.7 |
|   WISDMAtDataset | 71.6 | 28.4 |
|   REALDISPDataset | 100.0 | 0.0 |
|   HHARDataset | 61.3 | 38.7 |
|   MHEALTHDataset | 57.1 | 42.9 |

**Supplementary Table S2**. Number of samples used for training, validation, and evaluation.

| Dataset | Training | Validation | Evaluation |
|---|---|---|---|
| GaitLab | | | |
|     Number of HC samples | 52,212 | 2,906 | 64,932 |
|     Number of PwMS samples | 148,666 | 26,112 | 191,486 |
| Internal Roche dataset | | | |
|     Mean number of HC samples[a] | 32,160 | 6,432 | 3,216 |
| Publicly available data sources | | | |
|     Number HC samples | 41,856 | 18,028 | 0 |

[a]Due to the smaller sample size of the internal Roche dataset, a leave-one-out cross-validation approach was chosen to split the data across the training, validation, and evaluation sets, resulting in all the data being used in all three sets. For each fold (i.e., for each participant assigned to the evaluation set), data from two randomly chosen participants were assigned to the training set, and the data from all remaining participants to the validation set. Given this approach, only the average number of samples used in each of three sets are reported.

**Supplementary Table S3**. Number of subjects from GaitLab dataset used in the analysis for the detection on gait versus non-gait activities by walking aid usage status.

| Dataset | Sample size | |
| --- | --- | --- |
| | Without walking aid | With walking aid |
| PwMS with EDSS 0.0 – 3.5 | 19 | 1 |
| PwMS with EDSS 4.0 – 5.5 | 15 | 7 |
| PwMS with EDSS 6.0 – 6.5 | 3 | 23 |

EDSS, Expanded Disability Status Scale; PwMS, people with multiple sclerosis.

**Supplementary Table S4**. Number of subjects from GaitLab dataset used in the analysis on ambulatory gait bout duration versus MS disease severity.

| Dataset | Sample size | |
| --- | --- | --- |
| | Without walking aid | With walking aid |
| PwMS with EDSS 0.0 – 3.5 | 17 | 1 |
| PwMS with EDSS 4.0 – 5.5 | 14 | 7 |
| PwMS with EDSS 6.0 – 6.5 | 1 | 23 |

EDSS, Expanded Disability Status Scale; PwMS, people with multiple sclerosis.